\documentclass[a4paper,fleqn]{cas-sc}
\usepackage[numbers]{natbib}
\usepackage{amssymb}
\usepackage{amsmath}
\usepackage{graphicx}
\usepackage[linesnumbered, ruled]{algorithm2e}
\usepackage{booktabs}
\usepackage{makecell}
\usepackage{caption}
\usepackage{tabularx}
\usepackage{amsthm}
\usepackage{amstext}
\theoremstyle{definition}
\usepackage{float}
\usepackage{subfigure}
\usepackage{hyperref}
\DeclareUnicodeCharacter{202F}{FIX ME!!!!}
\def\tsc#1{\csdef{#1}{\textsc{\lowercase{#1}}\xspace}}
\tsc{WGM}
\tsc{QE}
\tsc{EP}
\tsc{PMS}
\tsc{BEC}
\tsc{DE}
\graphicspath{{figs/}}


\begin{document}
	\begin{sloppypar}
		\let\WriteBookmarks\relax
		\def\floatpagepagefraction{1}
		\def\textpagefraction{.001}
		
		\shorttitle{}
		
		\shortauthors{Jiahao Zhang et~al.}
		
		\title [mode = title]{Weakly supervised alignment and registration of MR-CT for cervical cancer radiotherapy}                      
		
		
		%
		\author[1]{Jiahao Zhang }[type=editor]
		\ead{ZJH302524@gmail.com}
		
		\author[1]{Yin Gu}[%
		style=chinese
		]
		\ead{guyinn96@gmail.com}
		
		\author[2]{Deyu Sun}[%
		style=chinese
		]
		\ead{91111@126.com}
		
		\author[3]{Yuhua Gao}[%
		style=chinese
		]
		\ead{52437438@qq.com}
		
		\author[2]{Ming Cui}[%
		style=chinese
		]
		\ead{17740059400@163.com}
		
		\author[2]{Teng Zhang}[%
		style=chinese
		]
		\ead{904828678@qq.com}
		
		\author[1,4]{He Ma}[
		style=chinese,
		orcid=0000-0002-5054-3586]
		\cormark[1]
		\ead{mahe@bmie.neu.edu.cn}
		
		\affiliation[1]{
			organization={College of Medicine and
				Biological Information Engineering},
			addressline={Northeasten University},
			city={Shenyang},
			postcode={110169}, 
			state={Liaoning},
			country={China}}
		\affiliation[2]{
			organization={Department of Radiation Oncology Gastrointestinal and Urinary and Musculoskeletal Cancer},
			addressline={Cancer Hospital of Dalian University of Technology},
			city={ Shenyang},
			postcode={110110}, 
			state={Liaoning},
			country={China}}
		\affiliation[3]{
			organization={Department of Gynecological Radiotherapy},
			addressline={Cancer Hospital of China Medical University},
			city={ Shenyang},
			postcode={110110}, 
			state={Liaoning},
			country={China}}
		\affiliation[4]{
			organization={Key Laboratory of Intelligent Computing in Medical Image},
			addressline={Ministry of Education},
			city={Shenyang},
			postcode={110819}, 
			country={China}}
		\cortext[cor1]{Corresponding author}
		
		\begin{abstract}	
			Cervical cancer is one of the leading causes of death in women, and brachytherapy is currently the primary treatment method. However, it is important to precisely define the extent of paracervical tissue invasion to improve cancer diagnosis and treatment options. The fusion of the information characteristics of both computed tomography (CT) and magnetic resonance imaging (MRI) modalities may be useful in achieving a precise outline of the extent of paracervical tissue invasion. Registration is the initial step in information fusion. However, when aligning multimodal images with varying depths, manual alignment is prone to large errors and is time-consuming. Furthermore, the variations in the size of the Region of Interest (ROI) and the shape of multimodal images pose a significant challenge for achieving accurate registration. In this paper, we propose a preliminary spatial alignment algorithm and a weakly supervised multimodal registration network. The spatial position alignment algorithm efficiently utilizes the limited annotation information in the two modal images provided by the doctor to automatically align multimodal images with varying depths. By utilizing aligned multimodal images for weakly supervised registration and incorporating pyramidal features and cost volume to estimate the optical flow, the results indicate that the proposed method outperforms traditional volume rendering alignment methods and registration networks in various evaluation metrics. This demonstrates the effectiveness of our model in multimodal image registration. 		
		\end{abstract}
		
		
		

		\maketitle
		
		\section{Introduction}
		
		Cervical cancer is one of the leading causes of cancer-related deaths among women worldwide, affecting nearly one million women each year~\cite{Arbyn_Weiderpass}. Currently, brachytherapy plays a crucial role in the treatment of cervical cancer. However, the current clinical challenge lies in accurately defining the extent of invasion of the paracervical tissues. A precise definition of the extent of invasion ensures effective irradiation of the cancerous cells, reduces damage to normal tissues, and allows for appropriate allocation of the irradiation dose. CT and MR images are the most commonly used imaging modalities for diagnosing and defining treatment methods for cervical region cancers. CT images exhibit a high level of contrast in the bone structure, allowing for clear expression of anatomical information. On the other hand, MR images have high contrast for soft tissues, enabling physicians to observe and analyze pathological changes in these tissues effectively. Therefore, integrating information from the two modalities is advantageous for enhancing diagnostic results and treatment planning, as it enables a precise delineation of the extent of paracervical tissue invasion and facilitates personalized treatment decisions~\cite{Jian_Azampour}.
		
		Due to the multimodal nature of images, different modal images may have varying image depths, slice gaps, and sampling intervals. Therefore, it is often necessary to preprocess and manually align the spatial positions of multimodal image pairs to ensure they have the same image depths before alignment. We found that most studies have not provided detailed descriptions of the image alignment process. Often, manual alignment of the spatial positions of the two images is necessary, which can be cumbersome. Alternatively, many of the original images required for alignment already have the same depth, and the final alignment is performed directly. The commonly used manual selection and subjective judgment methods require consideration of the layer spacing of multimodal images, image similarity, and other factors, which necessitate prior knowledge and experience about the anatomical structure of the images. This often results in significant errors and time-consuming manual alignment.

		As a fundamental step in image fusion, image registration aims to find the optimal transformation that best aligns the underlying anatomical structures  to achieve specific practical goals~\cite{Wu_He_Li_Zhu_Wang_Burstein}. Traditional registration methods, including ELASTIX~\cite{Klein_Staring_Murphy}, HAMMER~\cite{Shen_2007}, and Optical flow~\cite{Yang_El}, optimize the loss function through continuous iteration and apply image processing techniques such as keypoint detection, edge extraction, and region segmentation~\cite{Liu_Li,Xiao_Zijie,Yu_Ye_Gao}, etc. to maximize the similarity between fixed and moving images and improve the smoothness of the deformation field. Unfortunately, solving such optimization problems typically results in unsatisfactory computational efficiency and registration accuracy~\cite{He_He_Cao_2023}. Iterative solutions are inherently time-consuming and impractical for many clinical applications, particularly intraoperative procedures.
		
		In recent years, significant progress has been achieved in deep learning-based methods, including supervised, semi-supervised, and unsupervised approaches. During the training process, the input consists of source and target images. Features are extracted, and a deep convolutional neural network is utilized to predict the spatial transform field, which is then used to distort the moving image towards the target. The transform parameters are iteratively optimized during training, significantly enhancing the speed and accuracy of the registration.
		
		In this paper, we propose a spatial position alignment framework and a weakly supervised registration model. The geometric invariance of the skeletal structure in multimodal images can be leveraged for spatial position alignment. The spatial position alignment algorithm takes into account the layer gap relationship between different modalities and the similarity of the skeletal structure in the cross-sectional images. It selects the image with the least depth as the base to obtain a matching image with the same depth. The aligned images are used to train TransFlow, a hybrid model based on transformer and CNN. The model fully considers the cost volume of the optical flow in different layers of the pyramid and constantly utilizes contextual information to refine the optical flow field~\cite{Sun_Yang_}.
		
		The primary contributions of our work are:
		
		(1) A new similarity computation method based on connected domains proposed. This method takes into account the shape similarity of all the bone structures in the images, using it  for noise removal and optimization of labeled images.
		
		(2) An adaptive spatial alignment framework for multimodal image sequences is proposed. This framework can spatially align two multimodal images with different depths and achieve positional calibration.
		
		(3) A cost-volume multimodal registration network called TransFlow is proposed to efficiently align multimodal images and synthesize clear and informative multimodal fusion results. This aims to guide the subsequent diagnosis of cervical cancer.
		
		\section{Realted works}
		\subsection{Deformable image registration}
		Deformable image registration is often formulated as an optimization problem that utilizes pixel displacement fields to represent spatial transformations and measure the similarity between a moved image and a fixed image.  We implement the classical deformable image registration method using a moving image and a fixed image by solving the following optimization problem:
		\begin{equation}
			\phi^*=\underset{\phi}{\operatorname{argmin}}~L_{\text {sim }}(\phi(I_{m}), I_{f})+L_{r e g}(\phi)
		\end{equation}
		Where $\phi^*$ represents the optimal registration field that transforms the input moving image into a fixed image, $L_{\text {sim }}$ is a similarity function that quantifies the resemblance between the moved image and the fixed image, and $I_{m}$ and $I_{f}$ denote the moving image (source image) and the fixed image (target image). The $L_{r e g}$ is the regularization term penalty that constrains the deformation field. We will distort the moving image to estimate the deformation field $\phi$, and in particular, when we impose constraints on the deformation field $\phi$, we can make the deformation mapping microscopic and invertible, and preserve the topology~\cite{Vercauteren_Pennec}.
		
		In the field of Convolutional Neural Networks (ConvNets), Aria et al. proposed a traditional optical flow constraint that generates a dense motion field in the output layer ~\cite{Ahmadi_Patras_2016}. Guha utilized an optical flow constraint to introduce a combination of spatial transform networks and U-Net-like networks, showcasing the effectiveness of the VoxMorph network for brain image registration~\cite{Lee_Oktay_Schuh_}. Boah found that VoxelMorph, in conjunction with a DDPM, can generate deformed images, enabling continuous trajectory registration of the brain to the face~\cite{Kim_Han_Ye}.	Jian et al. introduced three conditional penalties (AC, OC, PC) to VoxelMorph-based non-rigid registration to address the registration of rigid objects~\cite{staring2007rigidity}. This was done to examine the volume loss before and after registering spine MR and CT images~\cite{jian2022weakly}. In addition, Bob et al. proposed DIRNet~\cite{de2017end}, a deep learning network that can be utilized for variational image registration. Then Mohammad et al. applied this research to the field of cervical cancer and investigated the impact of various versions of DIRNET for HDR brachytherapy~\cite{salehi2023deep}. Jingfan found that  adversarial training can be used to achieve  unsupervised registration and successfully achieved multimodal registration of MR and CT images in the brain and pelvis~\cite{fan2019adversarial}. Matthew proposed a novel, generalized framework (ISTN) that can be used to achieve MR brain image registration by utilizing information from structures of interest~\cite{lee2019image}. Wang combined deformable image registration with affine transformations in a constrained affine transform network structure that can be  used for multimodal MR image registration~\cite{wang2021multimodal}, By utilizing a weakly-supervised, label-driven formulation, Hu et al. achieved multimodal registration of MR images and ultrasound images of rectal and prostate sites~\cite{hu2018label}~\cite{hu2018weakly}.
		
		Furthermore, due to the transformer's larger receptive field, Dosovitskiy et al. applied it to images in NLP~\cite{Dosovits0} and made significant advancements in image recognition. Afterward, Swin Transformer and its variants demonstrated superior performance in target detection and semantic segmentation. Therefore, the transformer can also be a powerful tool for image registration. Chen et al. drew inspiration from vision-transformer (VIT)-based segmentation methods and combined VIT with ConvNets to enhance the performance of image registration ~\cite{Chen_He_}. In addition, the Swin-transformer is also integrated with ConvNet to introduce TransMorph, as well as diffeomorphic and Bayesian variants, for topology preservation and deformation uncertainty estimation. This model significantly enhances the performance of medical image registration ~\cite{Chen_Frey}.
		
		\subsection{Correlation layer}		
		In order to extract features from the patches of the two images, obtain the feature representation of the two images, and implement a method similar to standard matching, we introduce a "correlation layer" to calculate the relationship between the patches. The feature maps of the moved image and the fixed image are compared by multiplying the patches ~\cite{Sun_Yang_}, and then, for a square block of size $k:=2k+1$, the "correlation" of the two blocks centered on $x_1$ in the first mapping and $x_2$ in the second mapping is defined as follows:	
		\begin{equation}
			\label{corr}
			c\left(x_1, x_2\right)=\sum_{o \in[-k, k] \times[-k, k]}\left\langle I_1\left(x_1+o\right), I_2\left(x_2+o\right)\right\rangle
		\end{equation}
		Where $~I_{1}~$ and $~I_{2}~$ represent the individual channel features in the multi-channel feature maps of the two images, Eq.(\ref{corr}) differs from the neural network in which the data is convolved with the filter. Instead, the data is convolved with the data, and therefore, there are no training weights.
		Since computing all features makes forward and backward propagation difficult, the maximum displacement for comparison is limited here, and striding is introduced to simplify the computation and reduce the amount of computation.
		
		Given the maximum displacement $~b~$, and then compute the correlation $~c\left(x_1,x_2\right)~$ for each position $~x_{1}~$within the range of $~x_{2}~$ in a neighborhood of size $~D:=2d+1~$. In addition, steps s1 and s2 are used to globally quantize $~x_{1}~$ and to quantize  $~x_{2}~$ in a neighborhood centered on $~x_{1}~$. According to \cite{Dosovitskiy_Fischer_}, the size of the correlation result is a 4-dimensional tensor, and it produces a correlation value for every combination of two 2D positions,  i.e., the scalar product of the two vectors.
		
		\section{Method}
		\label{sec:meth}
		\subsection{OffsetCorrection}
		Before registering, the image is corrected for offset. The location of ROI of the multimodal image data varies in space, causing different offsets and distortions that can significantly affect registration when the data is restricted. After applying offset correction, the location of ROI is repositioned to the center of the image. The same operation is then performed on the labeled image, resetting the image size.
		\subsection{Connected Domain Similarity Measure}
		The Dice Similarity Coefficient (DSC) is commonly used to address errors in weakly supervised registration tasks. By definition, DSC cannot utilize the shape information for segmenting structures. The complete global context information is not taken into account, and therefore, it may not be relevant for anatomical plausibility. When we applied the spatial position alignment algorithm to measure the similarity between two labels of different modalities, using the DSC directly would result in significant errors due to the unique nature of the ROIs' locations in the two multimodal images and the bone label locations. To tackle this issue, we employed a DSC-based method that takes into account the number of connected domains in the labeled images, as well as the area and  aspect ratio, to calculate the similarity metric.
		\begin{equation}
			\label{sim}
			\mathcal{SIM}=\frac{1}{n} \sum_{i=1}^{n}\frac{\left|\left(I_{MR} \circ T[S_{\text {con-MR}}^{i, h*w}]\right) \cap\left(I_{CT} \circ  T[S_{\text {con-CT}}^{i, h*w}]\right)\right|}{\left|I_{MR} \circ T[S_{\text {con-MR}}^{i, h*w}]\right|+\left|I_{CT} \circ  T[S_{\text {con-CT}}^{i, h*w}]\right|}
		\end{equation}
		This function also needs to satisfy
		\begin{equation}
			\label{condtion}
			\left|\left[ h_{CT}/w_{CT}-h_{MR}/w_{MR} \right]\right|<\gamma
		\end{equation}
		Where $~h~$ and $~w~$ represent the length and width of a connected domain. The notation $~\left|\left[\right]\right|~$ indicates that the result is rounded to the nearest whole number and then to its absolute value. For example, In Eq. (\ref{sim}), we define $~I_{MR}~$ and $~I_{CT}~$ as the labeled images of the MR and CT images, and $~S_{\text{con-MR}}^{i, h*w}~$ and $~S_{\text{con-CT}}^{i,h*w}~$ as the connectivity domains in the current MR or CT image sorted based on the area of the rectangular box of the connectivity domains. The area is represented by the product of the length and width of the rectangular box, i.e., $~h \times w~$. $S~$ represents the outcome after sorting all connected domains based on their area size, from the largest to the smallest. Meanwhile, $~i~$ denotes selecting the ith connected domain after sorting. Let $~T~$ represent the current image transformation, and let $\circ$ denote the application of a specific transformation to the image. Here, $~\gamma~$ signifies the similarity limit of the connectivity domains of two different modalities. The connectivity domains we aim to obtain should be smaller than this limit. Additionally, we require that the number of connectivity domains of the two images, denoted as $~n~$, is the same. Furthermore, the similarity coefficients of all computed connectivity domains are averaged to obtain the similarity of the two images with different modalities. 
		
		\subsection{Spatical Position Alignment}\label{alig}
		\begin{figure*}[htbp]
			\vspace{-3mm}
			\centering
			\includegraphics[width=0.9\linewidth]{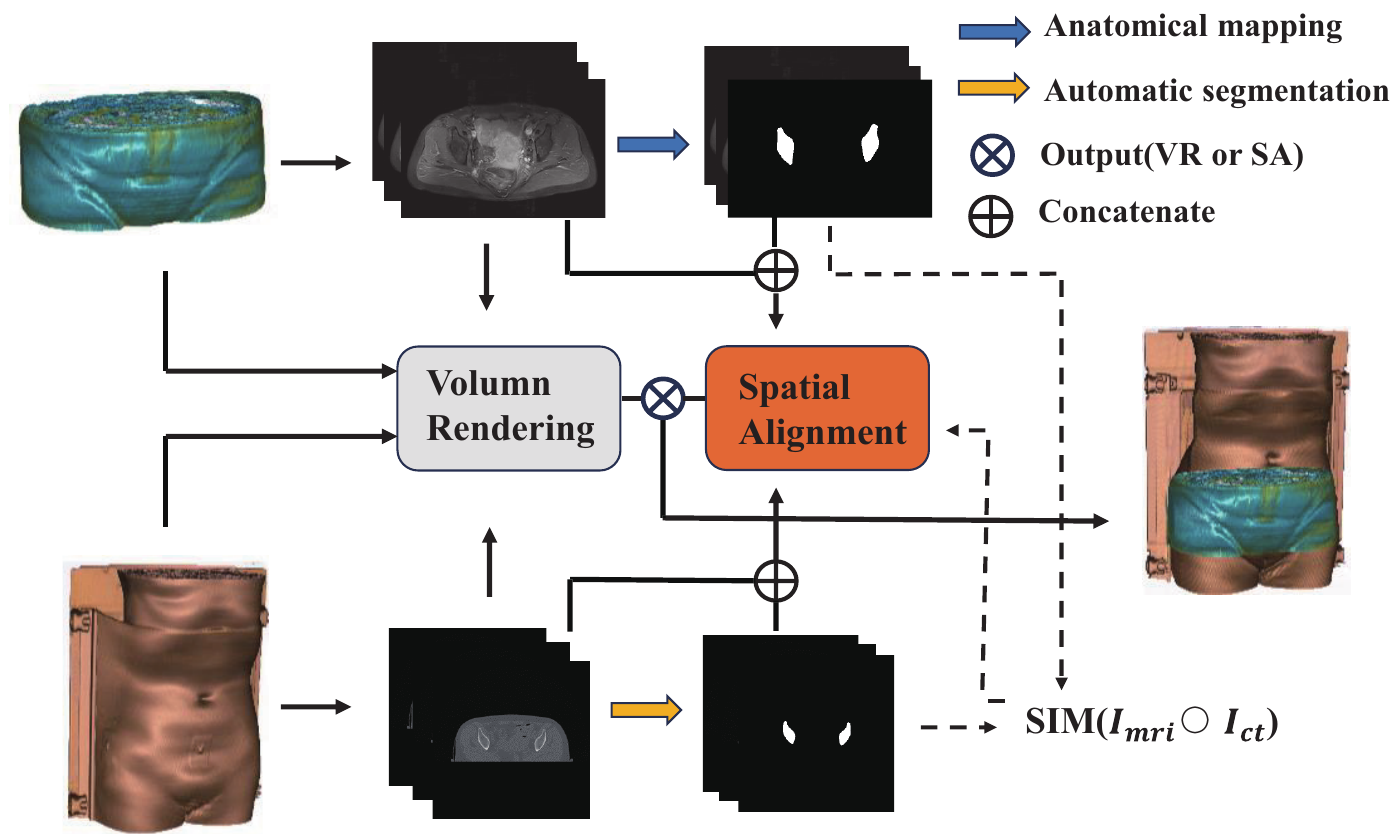}
			\caption{Spatial Position Alignment Steps. Volume Rendering: 3D Slicer using volume rendering for alignment; Spatial Alignment: Spatial position alignment algorithm using the original image and labeled image for automatic alignment.}\label{volume}
			\vspace{-3mm}
		\end{figure*}
		In medical imaging, we typically use layer gap and the number of layers to describe the resolution of voxels and the coverage of the image. Since images from different modalities often have varying layer gapes and numbers of slices, it may be inaccurate to directly match the image slices of two modalities one by one. A common approach involves using clinical markers or reference images for alignment. Typically, the visual features of anatomical bodies are utilized to align images from different modalities, often through manual selection of corresponding slices or subjective judgment, such as using volume rendering for manual alignment. In this study, we propose an algorithm for spatially aligning multimodal images using a slice mask of the pre-existing binary bone structure (e.g., Fig. \ref{volume}). This approach can automatically match the layer gap of the image with the anatomical image features to find the best correspondence.
		
		Due to variations in layer gap among different patients,  we acquire  images with layer gap information for both modalities of a person. The MR images containing bone masks (only partially drawn by the doctor) are then combined with all CT images to calculate the  similarity scores ($\mathcal{SIM}$) ( Eq. (\ref{sim}) ). The CT layer with the highest similarity is then combined with the MR layer to determine the best-matched layer $~\mathcal{BP}~$. Each MR layer has a best matching layer, and we begin with the MR-CT matching layer with the highest similarity. We filter the MR-CT counterparts that meet the layer gap formula in the set $~\mathcal{B}~$ of best matching layers $~\mathcal{BP}~$. The formula for layer gap  is :
		\begin{equation}
			\left\{
			\begin{aligned}
				\label{ceng}
				\mathcal{CT}_{k}^{compute}&=
				\left\{
				\begin{aligned}
					&\mathcal{CT}_{stand}^{j}-[|(\mathcal{MR}^{k}-\mathcal{MR}_{stand}^{j})|\\
					&\times\mathcal{GAP}_{MR}/\mathcal{GAP}_{CT}],k>j\\
					&\mathcal{CT}_{stand}^{j}+[|(\mathcal{MR}^{k}-\mathcal{MR}_{stand}^{j})|\\
					&\times\mathcal{GAP}_{MR}/\mathcal{GAP}_{CT}],k<j\\
				\end{aligned}
				\right.\\
				\mathcal{CT}_{Range}^{k}&=[\mathcal{CT}_{k}^{compute}-1,\mathcal{CT}_{k}^{compute},\mathcal{CT}_{k}^{compute}+1]\\
			\end{aligned}
			\right.
		\end{equation}
		where $~\mathcal{MR}_{stand}^{j}~$ and $~\mathcal{CT}_{stand}^{j}~$ represent the MR and CT image indices in the best matching layer $~\mathcal{BP}~$. $~\mathcal{MR}^{k}~$ denotes the  ordered set $~\mathcal{C}~$ of the  remaining MR image indexes, where $~k~$ and $~j~$ are indexes. The brackets $~[]~$ indicate rounding, $~||~$ denotes absolute value, $~\mathcal{GAP}_{MR}~$ and $~\mathcal{GAP}_{CT}~$ represent the layer gaps, and the computed CT image indexes, $~\mathcal{CT}_{k}^{compute}~$, are added to form the range of the set $~\mathcal{CT}_{Range}^{k}~$. The aim is to verify whether the $~\mathcal{MR}^{k}~$ corresponding to $~\mathcal{CT}^{k}~$ in the ordered set ensemble $~\mathcal{C}~$ is within the set $~\mathcal{CT}_{Range}^{k}~$, and to exclude  non-setting MR-CT correspondences to form a filter set $~\mathcal{EP}~$. Each MR layer corresponds to a filter set $~\mathcal{EP}~$, and the set $~\mathcal{EL}~$ with the longest length is extracted from the total set $~\mathcal{EPL}~$ formed by the filter set $~\mathcal{EP}~$,We then use ( Eq. ( \ref{ceng2} )) to optimize our filter.  Rounding is not used.  We use $\lfloor \rfloor$ for rounding down, $\lceil \rceil$ for rounding up, and $~\widehat{}~$ for getting the  fractional parts.  ( Eq. ( \ref{ceng2} )) is employed to blend the features of the front and back layers to enhance the accuracy of the matching process. Ultimately, we will obtain the set of MR-CT counterparts for all the anatomical labels, denoted as $~\mathcal{EL}~$. We describe the alignment process in the algorithm \ref{IPLR}.
		\begin{equation}
			\begin{aligned}
				\label{ceng2}
				\mathcal{CT}_{k}&=
				\left\{
				\begin{aligned}
					&\mathcal{CT}_{stand}^{j}-|(\mathcal{MR}^{k}-\mathcal{MR}_{stand}^{j})|\\
					&\times\mathcal{GAP}_{MR}/\mathcal{GAP}_{CT},k>j\\
					&\mathcal{CT}_{stand}^{j}+|(\mathcal{MR}^{k}-\mathcal{MR}_{stand}^{j})|\\
					&\times\mathcal{GAP}_{MR}/\mathcal{GAP}_{CT},k<j\\
				\end{aligned}
				\right.\\
				\mathcal{CT}_{\mathcal{CT}_{k}^{compute}}&=(1-\widehat{\mathcal{CT}_{k}})\times\mathcal{CT}_{\lceil\mathcal{CT}_{k}\rceil}\oplus \widehat{\mathcal{CT}_{k}}\times\mathcal{CT}_{\lfloor\mathcal{CT}_{k}\rfloor},
			\end{aligned}
		\end{equation}
		\begin{algorithm}
			\caption{Spatial Position Alignment }
			\label{IPLR}
			\KwIn{$\mathcal{CT}$\&$\mathcal{MR}$ pelvis image + bone labels}
			\KwOut{$\mathcal{EL}$}   
			\For{$j\leftarrow 1$ \textbf{to} $ \mathcal{MR}$}{
				\For{$i\leftarrow 1$ \textbf{to} $ \mathcal{CT}$}{
					Computed $\mathcal{SIM(CT_\textbf{i}, MR_\textbf{j})}$  according to Eqs. (\ref{sim}) and (\ref{condtion})\;}
				\textsc{\tcp{Find the CT for each layer of MR}}
				$\mathcal{BP}$$\leftarrow$ Select the best $\mathcal{SIM}$ value corresponding to the image pair $\mathcal{CT}_{stand}^{j}$, $\mathcal{MR}_{stand}^{j}$\;
				Copy best $\mathcal{SIM}$ to $\mathcal{A}$, $\mathcal{BP}$ to $\mathcal{B}$}
			\textsc{\tcp {Obtain the best MR-CT correspondence}}
			$\mathcal{A}_{sorted}^{l\rightarrow s}$, $\mathcal{B}_{sorted}^{l\rightarrow s}$$\leftarrow$Sort  $\mathcal{A}$ and  $\mathcal{B}$ from largest to smallest\;
			$\mathcal{C}_{sorted}^{\mathcal{MR}_{stand}^{j}}$$\leftarrow$Sort B according to $\mathcal{MR}_{stand}^{j}$ from  smallest to largest\;
			
			\While{$\mathcal{MR}_{stand}^{j}$$\nrightarrow$$\mathcal{C}_{sorted}^{\mathcal{MR}_{stand}^{j}}$.{end}}{
				$\mathcal{EP}$$\leftarrow$ Filter the Image pair sequence  using  $\mathcal{B}$ , Eqs. (\ref{sim}) and (\ref{ceng})\;
				Copy  $\mathcal{EP}$ to $\mathcal{EPL}$}
			\textsc{\tcp{Filter the best MR-CT correspondence}}
			$\mathcal{EL}_{MAX}^{\mathcal{EPL}}$$\leftarrow$The largest length in $\mathcal{EPL}$\;
			$\mathcal{EP}_{Index}^{\mathcal{EL}}$$\leftarrow$the maximum value index in $\mathcal{EPL}$\;
			Update $\mathcal{EL}$ using $\mathcal{EP}$ , Eqs. (\ref{sim}) and (\ref{ceng2}) \;
			\textsc{\tcp{Optimized MR-CT correspondence}}
		\end{algorithm}
		\vspace{-0.5cm}
		\subsection{Label denoising}
		\begin{figure*}[!h]
			\centering
			\subfigcapskip=-8pt
			\subfigure[]{
				\includegraphics[width=0.148\linewidth]{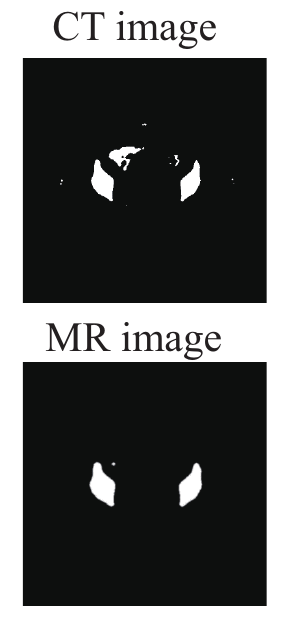}
			}\hspace{-3mm}
			\subfigure[]{
				\includegraphics[width=0.3\linewidth]{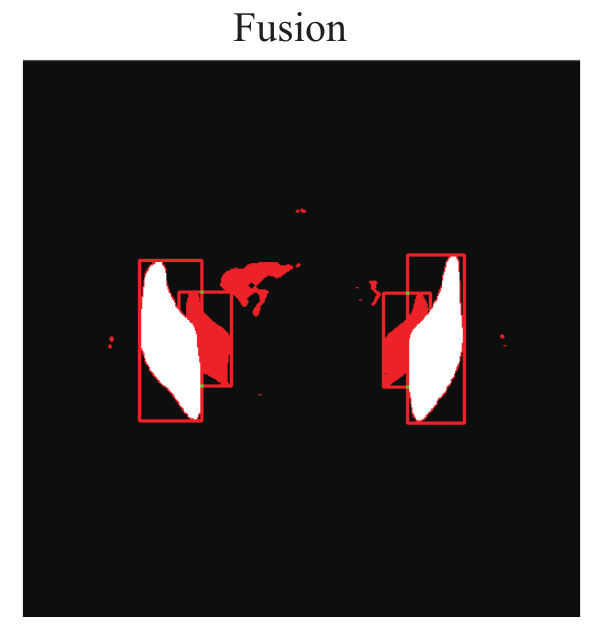}
			}\hspace{-3mm}
			\subfigure[]{
				\includegraphics[width=0.195\linewidth]{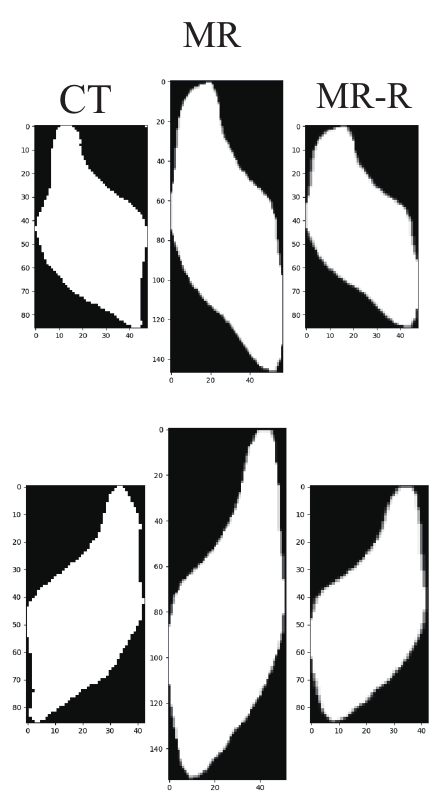}
			}\hspace{-3mm}
			\subfigure[]{
				\includegraphics[width=0.258\linewidth]{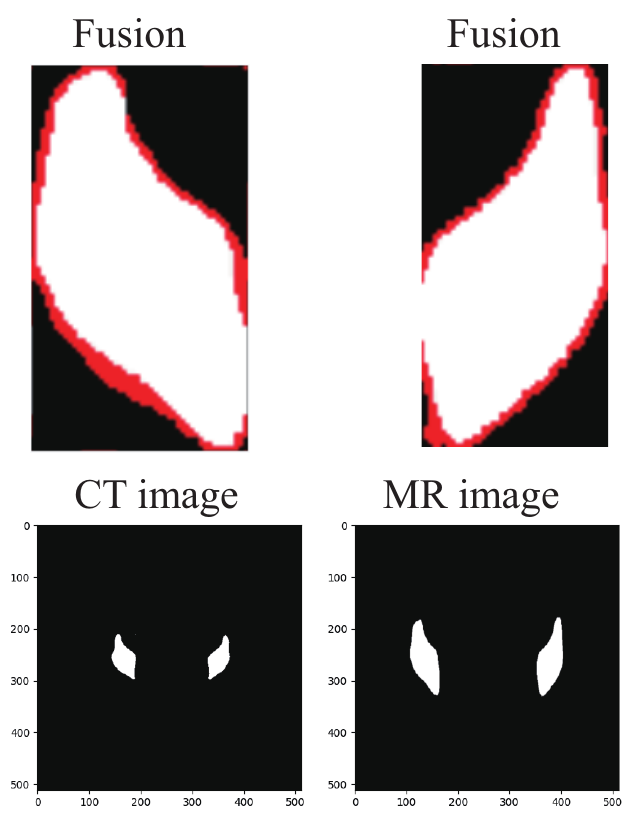}
			}\hspace{-3mm}
			\caption{The $\mathcal{SIM}$ function denoising process is shown from left to right. (a): The top image is the  CT labeled image before denoising, and the bottom image is the MR labeled image. (b) The range of attention of the $\mathcal{SIM}$ function. (c) progresses from left to right, including the local CT image, the local MR image, and the local MR image after resizing. (d) The top image shows the fused local labeled image, while the bottom image shows the denoised CT and MR labeled image.}\label{SIM progress}
			\vspace{-3mm}
		\end{figure*}
		We use Eq. (\ref{sim}) not only to calculate the similarity of two labeled images, but also for noise removal from the original roughly labeled images, Fig. \ref{SIM progress} shows in detail the denoising process of the $\mathcal{SIM}$ function,  the upper figure ( Fig. \ref{SIM progress}-a ) represents the CT labeled image obtained after automatic segmentation, while the lower figure ( Fig. \ref{SIM progress}-a )  represents the MR labeled image obtained by anatomical mapping. Firstly, the $\mathcal{SIM}$ function filters the connectivity domains in the two labeled images according to the $~\gamma~$ value in Eq. (\ref{condtion}) , and removes the connectivity domains that do not meet the requirements of Eq. (\ref{condtion}),  Fig. \ref{SIM progress}-b represents the connected domains of interest obtained according to Eq. (\ref{condtion}) , and Fig. \ref{SIM progress}-c represents the connected domains in CT, the connected domains in MR, and the connected domains of the same size resized by MR according to CT, from left to right.  The two images above Fig. \ref{SIM progress}-d represent the effect of superposition of the two images after transformation, and the similarity is determined by solving the overlap degree according to the transformed connectivity domain. The two images below Fig. \ref{SIM progress}-d after the denoising operation are finally obtained, and we can see that the noise of the images is removed, leaving only the corresponding similarity mask part.
		\subsection{Deep Neural Network Architecture}
		\begin{figure*}[htbp]
			\centering
			\includegraphics[width=0.9\linewidth]{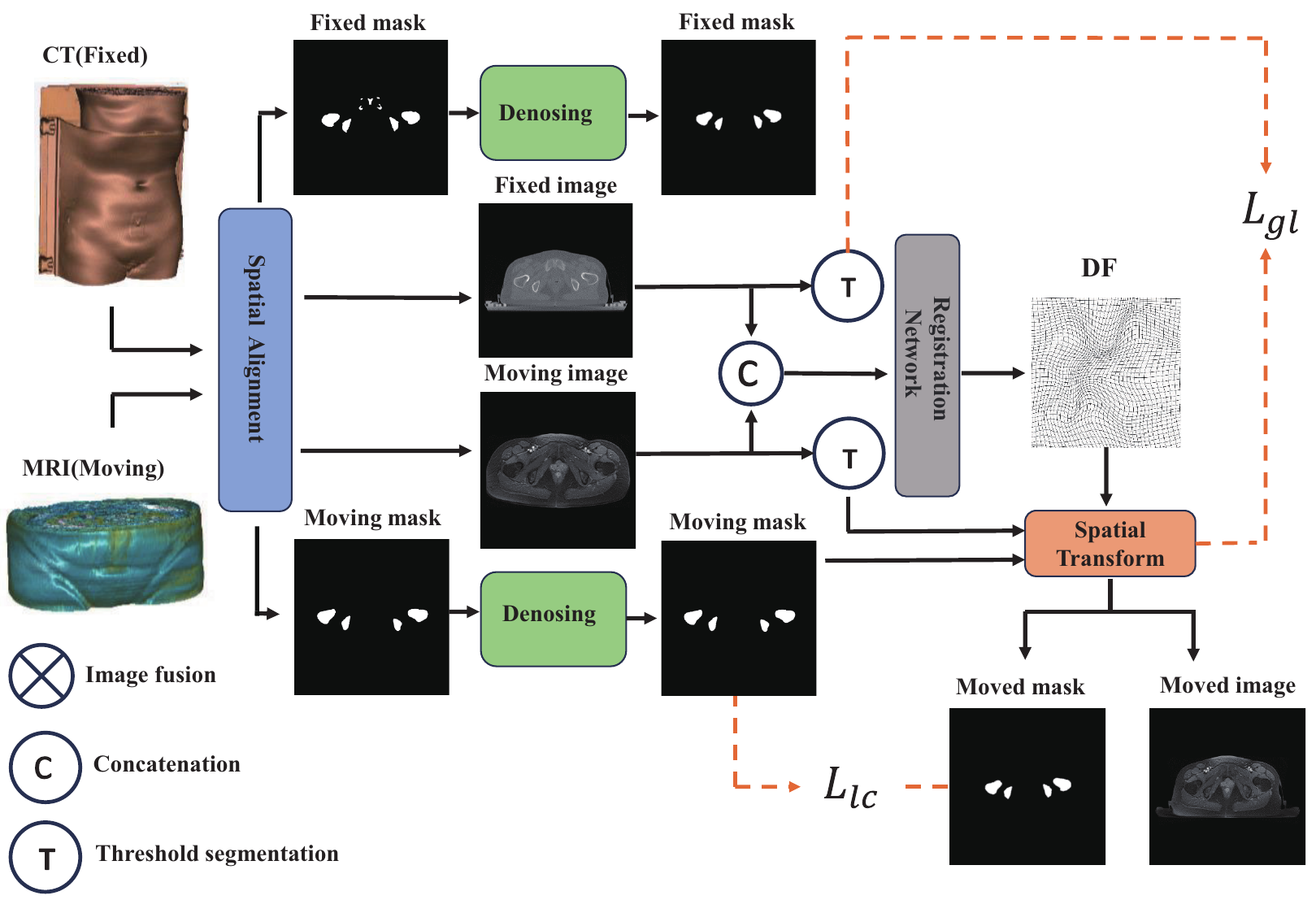}
			\caption{The overall process of multimodal registration is shown in the schematic diagram. Firstly, the multimodal images are spatially aligned for interlayer registration, then put into the network for training, the binary ROI contour masks and bone labels of the original image are transformed using the deformation field and the  loss is calculated.$L_{l c}$ and $L_{g l}$}\label{network}
			\vspace{-3mm}
		\end{figure*}
		\begin{figure*}[htbp]
			\centering
			\includegraphics[width=1\linewidth]{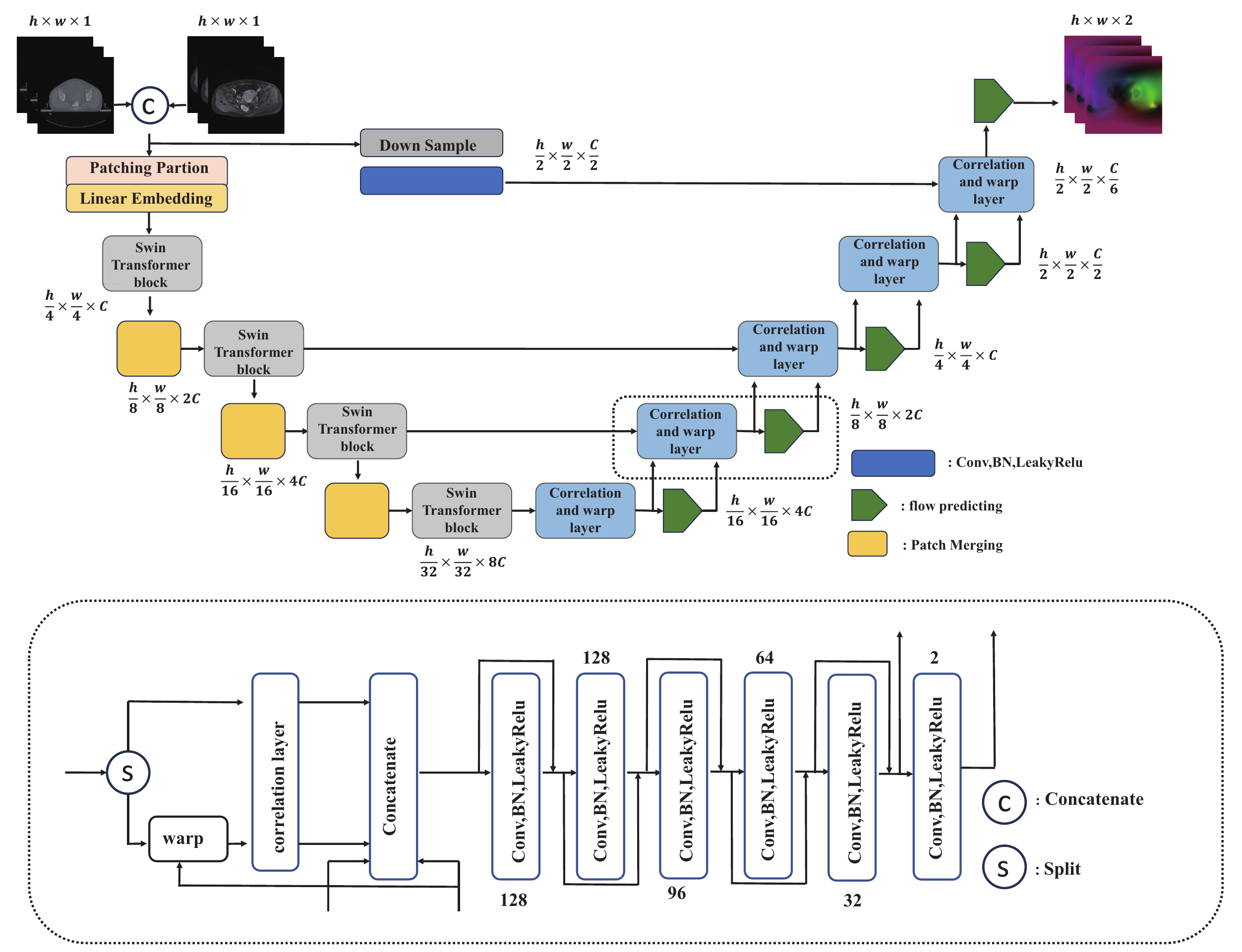}
			\caption{Overview of the registration network}\label{network2}
			\vspace{-3mm}
		\end{figure*}
		The overall registration process is depicted in Fig. ~\ref{network}, the image first undergoes spatial position registration using an algorithm to achieve inter-layer registration. It is then sent to the network for training. The similarity of the thresholded segmented binary ROI contour masks and the denoised bone labels are calculated respectively by the deformation field obtained. As shown in Fig. \ref{network2} , in the encoder part, we adopt a similar structure to the Transmorph~\cite{Chen_Frey}. After image splicing, the output resulting in a size of $~\frac{H}{2} \times \frac{W}{2} \times \frac{C}{2}~$ after the convolutional layer (as depicted in the blue box in Fig. ~\ref{network2}) to obtain a high-resolution feature map. After that, there are 4 levels of SwinTransformer modules and 3 levels of patch merging (as indicated by the yellow box in Fig. \ref{network2}) , the number of channels is expanded by a factor of two each time. As a result, the final output dimension of the encoder is  $~\frac{H}{32} \times \frac{W}{32} \times 8C~$. The decoder part utilizes  correlation layers, as indicated by the dashed box in Fig. \ref{network2}. In the network architecture, each input is divided into two images with an equal number of channels. The correlation between the transformed image and the other is then estimated using a correlation layer. This is followed by splicing with the output of the previous correlation-warp  layer, and  successive convolutional layers  are continuously and repeatedly spliced together. The final convolutional layer is referred to as the optical flow prediction layer, as depicted in the green graph in Fig. ~\ref{network2} . Multiple scales or multiple layers of optical flow prediction layers  use different weights, working together to learn the representation of image features and attempt to perform optical flow estimation at different scales. This enhances the generalization ability of the model, as the Transformer may not be able to provide  high-resolution feature maps and cannot aggregate local information at lower levels. ~In \cite{Raghu_Unter}, we  introduce a convolutional layer to capture local information and generate high-resolution feature maps spliced with the output of the previous scale. Finally, we generate the deformation field and subsequently use the spatial transformation function\cite{Jaderberg_Si} to obtain the transformed image.
		\subsection{Loss function}
		The overall loss function of the network comprises four components. The first component calculates the similarity between the moved image and the fixed image $~L_{lc}~$ ( Fig. \ref{network} ). The second component measures the degree of overlapping layers of the two binary ROI contour masks $~L_{gl}~$ ( Fig. \ref{network}). Finally, there is a regularization of the overall deformation field with respect to the bone deformation field: 
		\begin{equation}
			\begin{aligned}
				\mathcal{L}_{loss}=&\lambda_{1} L_{lc}\left(I_f, \epsilon \left(I_m\right)\right)+\lambda_{2} L_{gl}\left(I_{f-t}, \epsilon\left(I_{m-t}\right)\right)+\\
				&\lambda_{3} L_{df}\left(\epsilon\right)+\lambda_{4}L_{df}\left(\epsilon\times I_m\right)
			\end{aligned}
		\end{equation}
		where $~\lambda_{1-4}~$ represents the weight factor. The symbol $~\epsilon~$ represents the deformation field, while $~I_f~$ and $~I_m~$ stand for the fixed labeled image and the moving labeled image, respectively. Additionally, $~I_{f-t}~$ and $~I_{m-t}~$ refer to the ROI contour masks of the fixed image and the moving image. $~L_{df}~$ represents the field smoothness constraint, defined as the total variation of the displacement field. In this work, we use DSC as the image similarity measure function $~\mathcal{L}_{\text{sim}}~$ in the loss function.
		\begin{equation}
			\mathcal{L}_{\text{sim}}(I_f, \phi\left(I_m\right))=1-{\frac{2 \times\left|I_f \cap \phi\left(I_m\right)\right|+\sigma}{\left|I_f\right|+\left|\phi\left(I_m\right)\right|+\sigma}}\label{dice}
		\end{equation}
		To prevent numerical instability, a smoothing factor $~\sigma~$ is added  here. We apply a diffusion regularizer to the spatial gradient of the  displacement field $~u~$ to promote smoothing of the displacement field.
		\begin{equation}
			\begin{aligned}
				\mathcal{L}_{\text {df}}(\epsilon)=&\sum_{p\in \Omega}\|\nabla u(p)\|^2,\\
				\nabla u(p)=&\left(\frac{\partial u(p)}{\partial x}, \frac{\partial u(p)}{\partial y}, \frac{\partial u(p)}{\partial z}\right)
			\end{aligned}
		\end{equation}
		and using the difference between neighboring voxels to approximate the spatial gradient, For $\nabla u(p)$, the approximation is made by using $\frac{\partial u(p)}{\partial x} \approx \linebreak u\left(\left(p_x+1,  p_y, p_z\right)\right)-u\left(\left(p_x, p_y, p_z\right)\right)$, the same is true for $\frac{\partial u(p)}{\partial y}$ and $\frac{\partial u(p)}{\partial z}$.
		\section{Experiments}
		\subsection{ Datasets and evaluation metrics}
		The CT and MR images of cervical cancer in this paper were obtained from  Liaoning Provincial Tumor Hospital in China. The data included images from a total of 90 patients. The MR images were scanned during the patients' admission examinations, while the CT images were scanned before the patients underwent extracorporeal radiation therapy. Each patient's data set contained approximately 20 MR slices and 90-110 CT slices. The bone regions in the MR images were  manually delineated by an experienced radiation oncologist. Among all the images, the size and shape of the labels for MR and CT images are not exactly the same. We utilize a spatial location alignment algorithm to align the multimodal images in order to obtain the image pairs for training. The total number of available image pairs is 1406. We train using 5-fold cross-validation, with 400 iterations per fold, a learning rate of 0.001, and a batch size  set to 2. The values of $\lambda_{1-4}$ are set to 1, 4, 3, and 4, respectively.
		
		We evaluate our approach from two distinct perspectives. In the context of spatial location alignment algorithms, we utilize four similarity metrics: 1) our metric ($\mathcal{SIM}$) (Eq. (\ref{sim})), 2) Mutual Information (MI)~\cite{Viola_}, 3) Dice coefficient (DSC), 4) Normalized cross-correlation (NCC)~\cite{AVANTS}, where $\mathcal{SIM}$ and DSC  only consider the average similarity of the labeled images, while MI and NCC consider the average similarity of the original images.
		
		For the registration algorithm, we also use various metrics: 1) Dice coefficient (DSC), 2) Hausdorff distance (HD), and 3) structural similarity index (SSIM)~\cite{Chen_Li}, 4) Jacobian coefficient (JC), 5) Jacobi determinant (JD). We use DSC and JC to measure the overlap of labeled images before and after the transformation, and HD and SSIM to measure the contour similarity of the image's location of interest. We primarily focus on the overall contour of the original image and pay attention to the overall registration effect. At the same time, we also take into account the displacement field of the Jacobi determinant. The Jacobi determinant is used to measure the smoothness and consistency of the optical flow field.
		
		%
		%
		%
		%
		
		\subsection{Comparison of new similarity measurement methods}
		\begin{figure*}[htbp]
			\centering
			\includegraphics[width=1\linewidth]{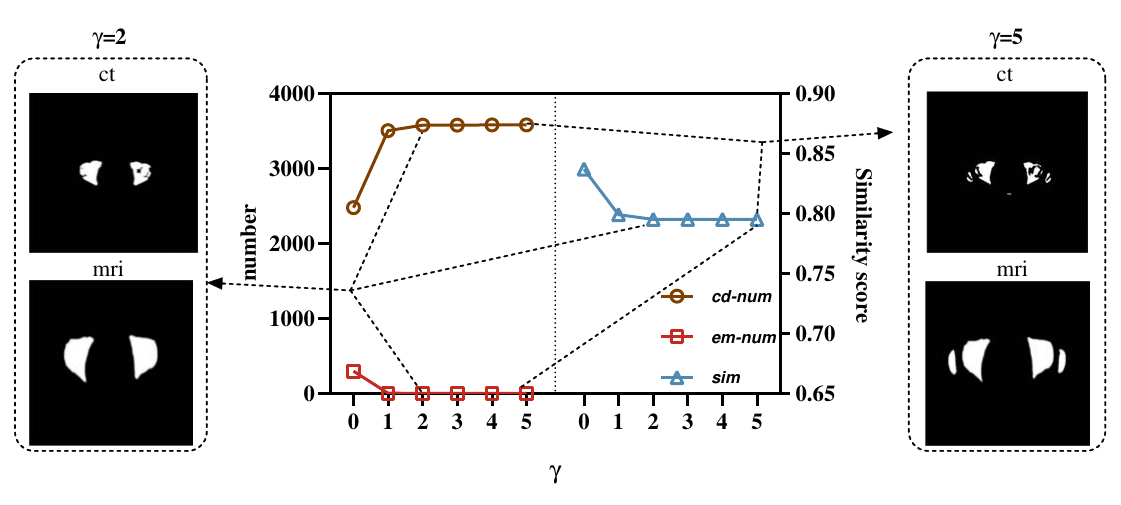}
			\caption{Average similarity ($\mathcal{SIM}$), connected domain retention ($\mathbf{cd-num}$) (for individual images), and empty image retention ($\mathbf{em-num}$) (for the dataset) trend with the $\gamma$ value. $\gamma=2$ exhibits less noise and rounder edges than $\gamma=5$.}\label{SIM}
			\vspace{-3mm}
		\end{figure*}		
		When denoising during the alignment process (see Fig. \ref{volume}) using Eqs. (\ref{sim}) and (\ref{condtion}), the magnitude of the $\gamma$ value is taken into account, and we examine the effect of different $\gamma$ values on the experimental results by comparing the average similarity ($\mathcal{SIM}$) ( Eq. (\ref{sim}) ), connected domain retention ($\mathbf{cd-num}$) (for individual images) and empty image retention ($\mathbf{em-num}$)(for the dataset), Fig. \ref{SIM} demonstrates the variation of the three variables with the value of $\gamma$. As the value of $\gamma$ decreases, it imposes greater restrictions on all connected domains in multimodal images, reduces the number of connected domains in labeled images, increases the number of empty images, and raises the average similarity $\mathcal{SIM}$. Conversely, as the value of $\gamma$ increases, it reduces the restriction on all connected domains in multimodal images, increases the number of connected domains in labeled images, decreases the number of empty images, but inevitably introduces noise. As shown in Fig. \ref{SIM}, a $\gamma$ of 2 produces less noise and a more rounded label compared to a $\gamma$ of 5, while a $\gamma$ of 0 leads to the complete disappearance of the connected domain of the image, resulting in a pure black mask.
		\subsection{Comparison of alignment algorithms }
		\begin{figure*}[htbp]
			\centering
			\subfigcapskip=-8pt
			\subfigure[]{
				\includegraphics[width=0.51\linewidth]{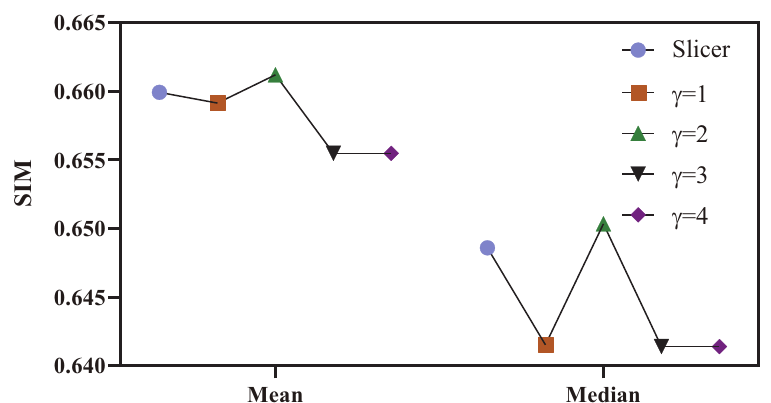}
			}\hspace{-3mm}
			\subfigure[]{
				\includegraphics[width=0.44\linewidth]{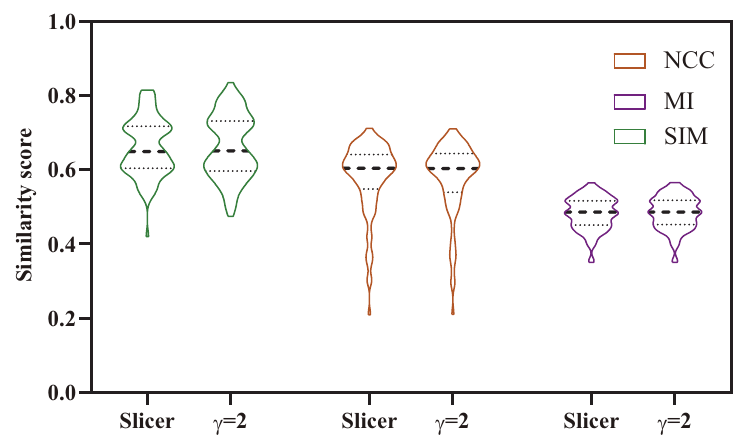}
			}\hspace{-3mm}
			\caption{(a):Results after averaging the inter-layer similarity for every one and then averaging the number of people again using different alignment methods, (b):Similarity distribution of 90 patients after averaging only the inter-layer similarity for each one using two alignment methods. }\label{meanandmedian}
			\vspace{-3mm}
		\end{figure*}
		We compared the spatial position alignment algorithm with the volume rendering spatial alignment method based on 3D Slicer. We utilized data from all patients, and all patients underwent the volume rendering and spatial position alignment algorithm to ensure that the MR and CT images were spatially aligned ( Fig. \ref{volume} ). For subsequent comparisons, all images were automatically offset-corrected by default. To assess the effectiveness of the volume-rendering and spatial alignment algorithms, we compared the similarity between CT  and MR images of all patients using different algorithms. We use $\mathcal{SIM}$ ( Eq. (\ref{sim}) ) to assess the performance of alignment algorithms, including various $\gamma$ alignment algorithms and the volume rendering alignment algorithm. We then calculated the average  similarity between all corresponding layers of all patients and the median, as shown in Fig. \ref{meanandmedian}-a, we can observe that when $\gamma$ is set to  2, it outperforms the other alignment scenarios, exhibiting the highest mean and median similarity. Additionally, the overall mean similarity of Slicer's volumn rendering is superior to all other scenarios except for when $\gamma$ is equal to 2.  
		
		We reexamined the distribution of alignment similarity across 90 patients by averaging all layers for each individual( Fig. \ref{meanandmedian}-b ). With $\gamma$ set to 2, the similarity distribution occurs more frequently at higher locations than the Slicer's volume rendering. Additionally, for the original image of MI compared to $\mathcal{SIM}$, the similarity does not produce much difference.
		\begin{figure*}[htbp]
			\centering
			\includegraphics[width=1\linewidth]{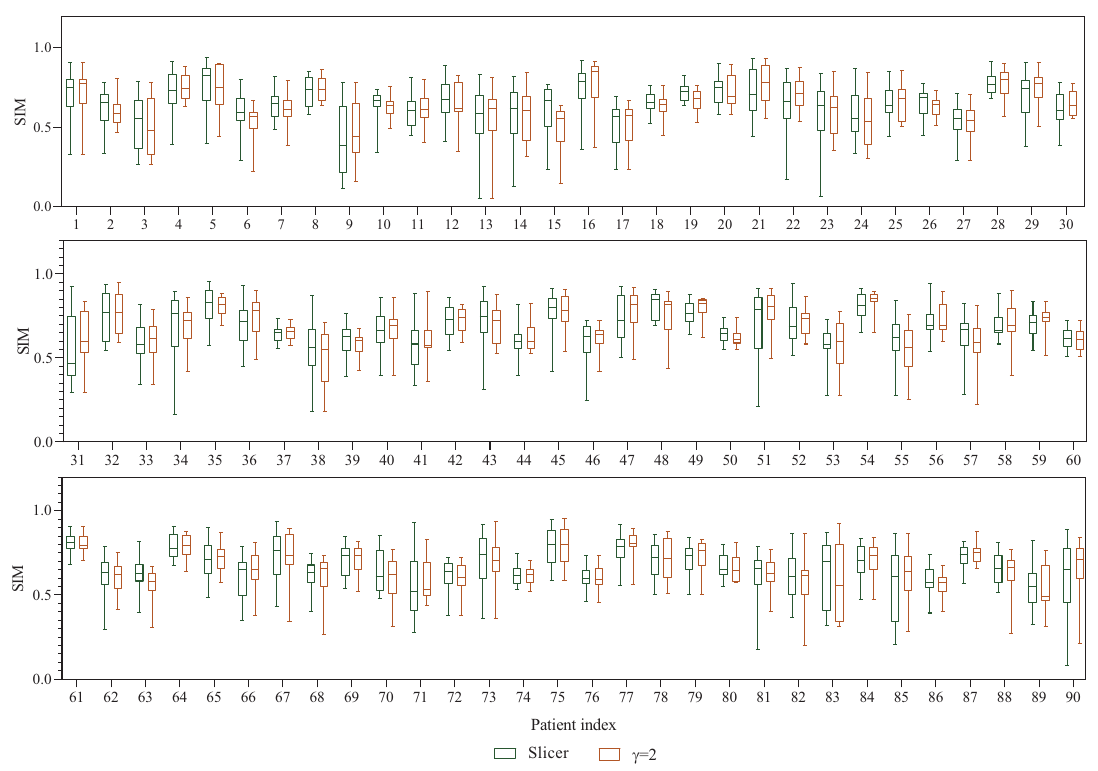}
			\caption{Distribution of inter-layer similarity per patient for different alignments}\label{align}
		\end{figure*}
		
\begin{figure*}[htb]
			\centering
			\includegraphics[width=1.02\linewidth]{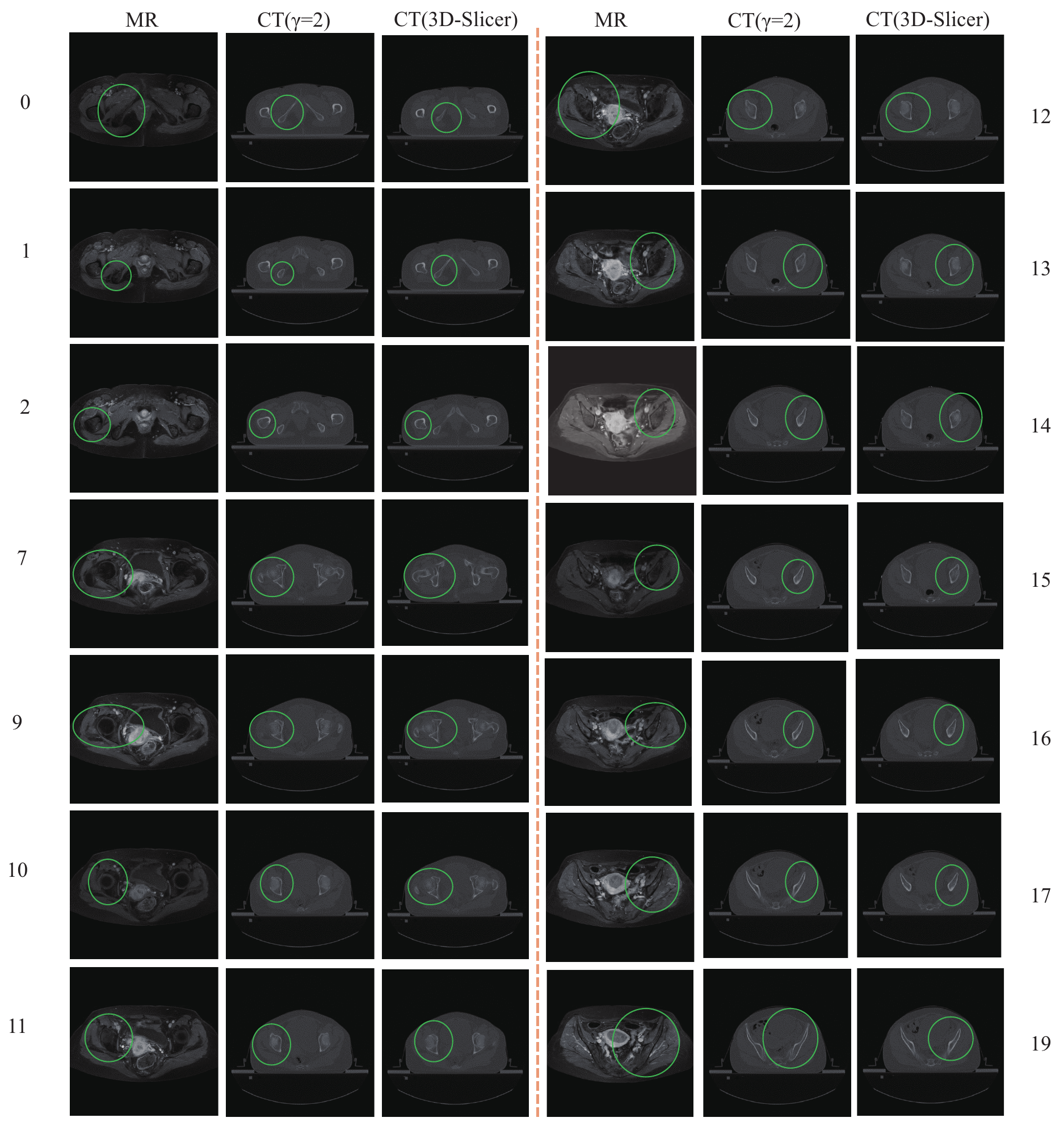}
			\caption{Results of the comparison between the spatial position alignment algorithm ($\gamma$=2) and the 3D Slicer volume rendering algorithm using MR images of patient 28 are shown below. Differences in effect are labeled using green \textcolor{green}{\textcircled{}}, and the leftmost and rightmost indexes represent the layers of the MR images , The remaining CT images depict the outcomes of the current MR layer matching.}\label{compair2}
			\vspace{-3mm}
		\end{figure*}

		We further examine the similarity distribution of all layers for all patients. Fig. \ref{align} provides a detailed illustration of the similarity distribution of inter-layer alignment for each of the 90 patients. For the most part, the overall distribution of similarity for spatial location alignment algorithms is better than or equal to that of volume rendering algorithms. Fig. \ref{compair2} provides a detailed comparison of the alignment effect, MR and CT using the volume-rendering  results obtained at the beginning of a large deviation from the correct results. The  volume rendering results from the 3D slicer for  bone assessment are not very accurate.
		Only manual alignment can achieve interlayer correspondence, but the  error is very large. Our approach considers the similarity of the contours and achieves more precise results by evaluating the contours of the bone images. Furthermore, we can directly obtain  aligned 2D slices  from the anatomical labels, which enhances efficiency and ensures  accuracy simultaneously.
		
		\subsection{Comparison of methods of registration }
		\begin{table*}[htbp]
			\caption{Comparison of various registration networks based on DSC, HD, SSIM, JC, $\mathcal{SIM}$, Std(JD), and \%|JD|<0}\label{compair-net}
			\setlength{\tabcolsep}{5.2mm}{
				\begin{tabular}{@{}p{2.4cm}lllllll}		
					\toprule
					Network & DSC~$\uparrow$ &HD~$\downarrow$& SSIM~$\uparrow$ &JC~$\uparrow$ &$\mathcal{SIM}$~$\uparrow$&std(JD)~$\downarrow$&\%|JD|<0~$\downarrow$ \\ \midrule
					ElasticSyN~\cite{Wetherall_Guttag} & 0.643 &11.367& 0.935&0.505&\quad&\quad&\quad \\
					VoxelMorph1~\cite{Balakrishnan} & 0.768&9.231& 0.952&0.659&0.738&0.354&0.009 \\
					VoxelMorph2~\cite{Balakrishnan} & 0.773 &8.870& 0.954&0.665&0.740&0.353&0.010 \\
					ACRegNet~\cite{Mansilla_Milone} & 0.794&9.234& 0.945&0.699&0.759&0.366&0.010 \\
					Vit-V-Net~\cite{Chen_He_} & 0.799&9.354& 0.960&0.703&0.764&\textbf{0.287}&0.006 \\
					TransMorph~\cite{Chen_Frey}& 0.805 &\textbf{8.183}& \textbf{0.967}&0.713&0.763&0.289&0.006 \\
					LKU-Net~\cite{Jia_Bartlett_Zhang} & 0.783&9.052& 0.962&0.681&0.760&0.308&0.006 \\
					PIViT~\cite{Ma_Dai_Zhang_Wen} & 0.780 &9.286& 0.951&0.676&0.765&\textbf{0.287}&\textbf{0.005} \\
					TransFLow & \textbf{0.825}&8.763& 0.959&\textbf{0.740}&\textbf{0.780}&0.295&0.006 \\\bottomrule
				\end{tabular}
			}
			\vspace{-3mm}
		\end{table*}
		
		Table ~\ref{compair-net} presents the results of network comparison for aligning MR images to CT images using traditional methods such as ElasticSyN \cite{Wetherall_Guttag}, VoxelMorph1 \cite{Balakrishnan}, VoxelMorph2 \cite{Balakrishnan}, ACRegNet \cite{Mansilla_Milone}, Vit-V-Net \cite{Chen_He_}, TransMorph \cite{Chen_Frey}, and LKU-Net \cite{Jia_Bartlett_Zhang}. PIViT \cite{Ma_Dai_Zhang_Wen} all achieved relatively low DSC, JC, and $\mathcal{SIM}$ scores compared to TranFlow. The best result was obtained by Transmorph, with  TransFLOW showing improvements of 2.5\% in DSC, 3.7\% in JC, and 2.1\% in $\mathcal{SIM}$ compared to Transmorph. This indicates that the transformer-based network structure outperforms the CNN structure in extracting multimodal features. However, TransFlow is slightly less effective than TransMorph in fitting the edges, with HD and SSIM scores 7\% and 0.8\% lower, respectively. Due to our comprehensive consideration of optical flow and local optical flow penalty, in addition to the traditional  method of ElasticSyN, the variance of the Jacobian determinant of the optical flow field in other networks ( std(JD) ) and the percentage of non-positive Jacobian determinant scores ( \%|JD|<0 ) have yielded excellent results.
		
		\begin{figure*}[htbp] 
			\centering
			\subfigcapskip=-8pt
			\subfigure[]{
				\includegraphics[height=0.56\linewidth]{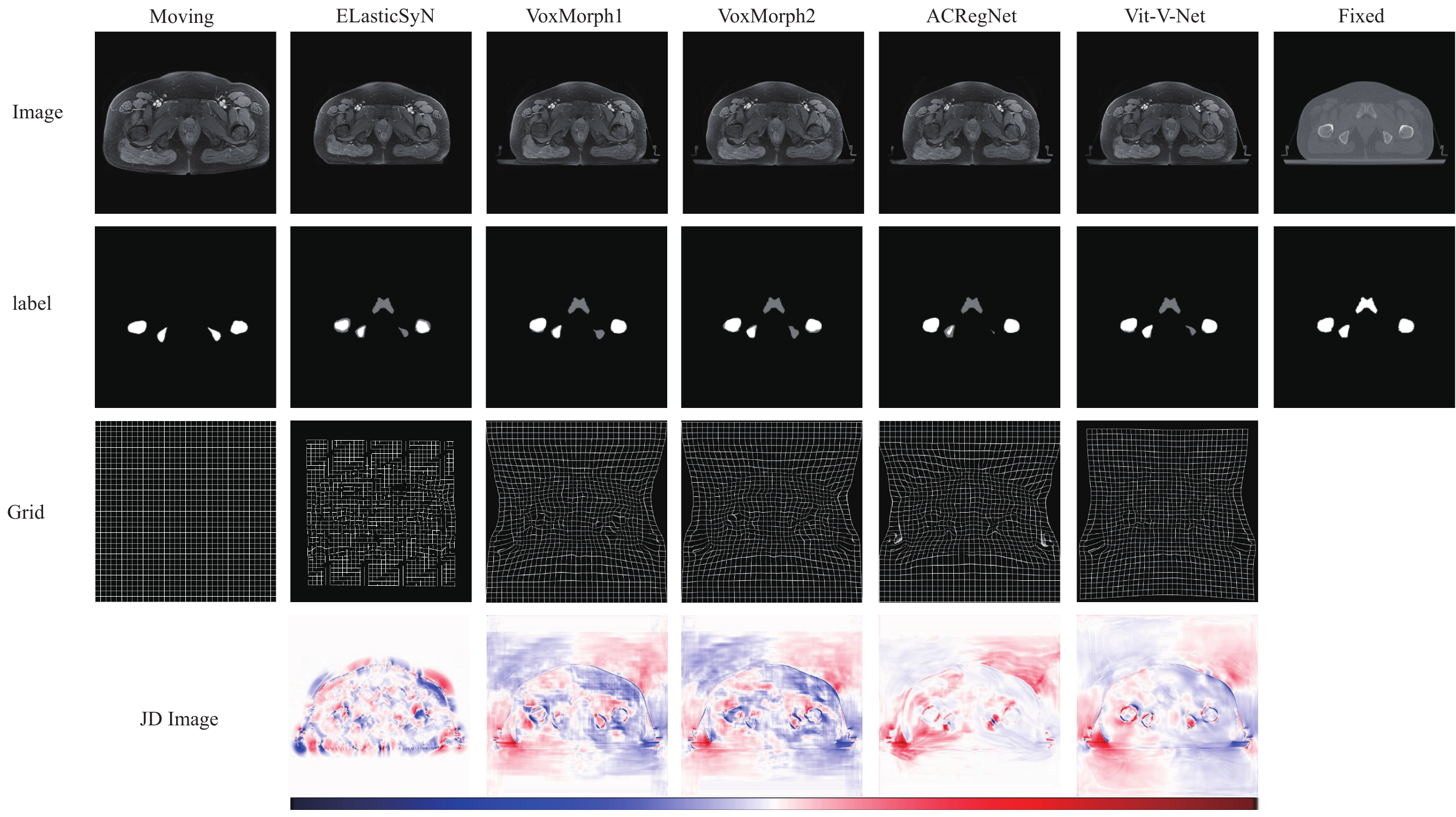} 
			}\hspace{5mm} 
			\subfigure[]{
				\includegraphics[height=0.65\linewidth]{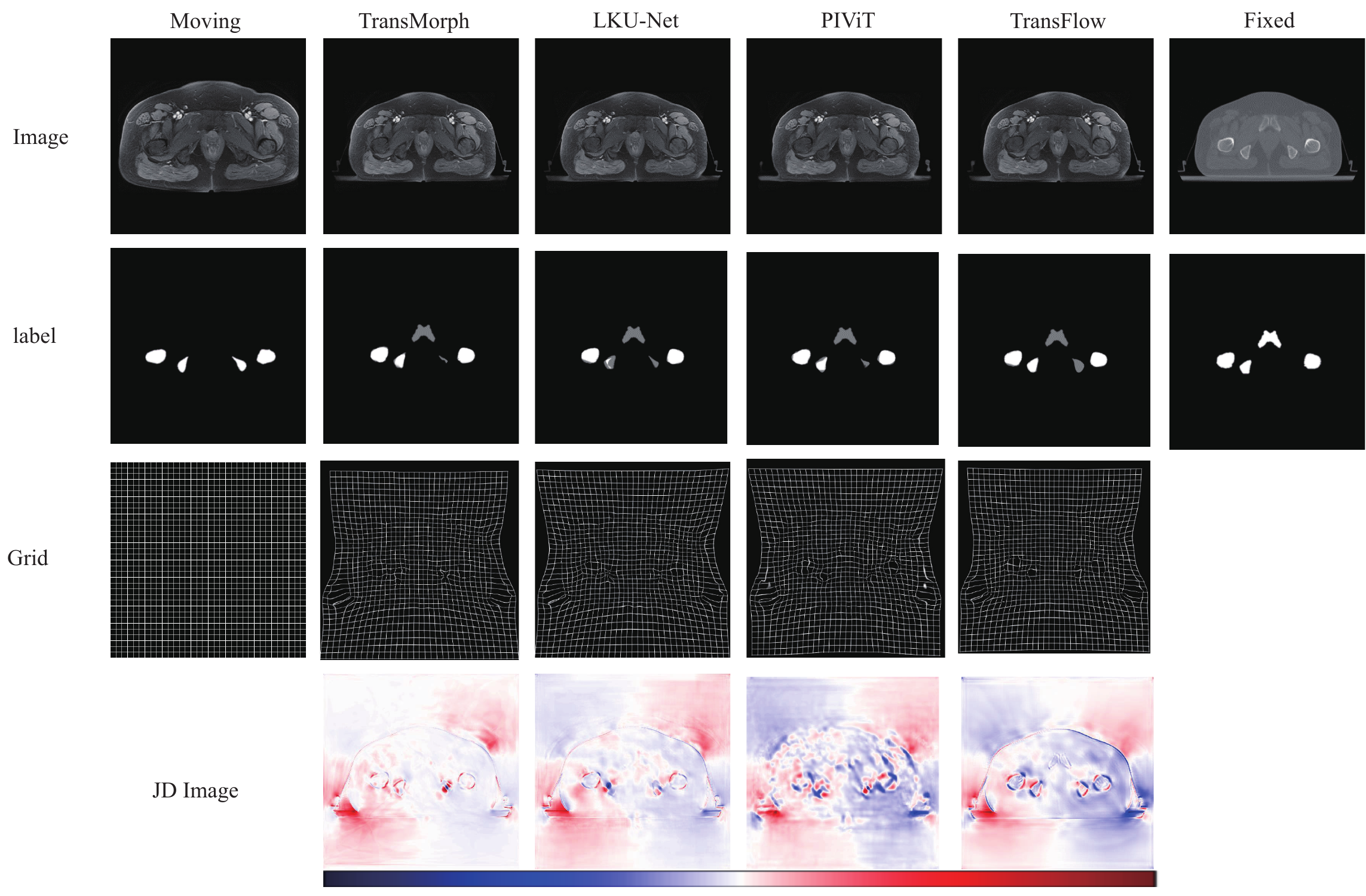} 
			}\hspace{5mm} 
			\caption{The comparison results for each network are presented in the following format: the first column from top to bottom shows the moving image, moving labeled image, and transformed field grid, while the last column displays the fixed image, fixed labeled image. The remaining columns showcase the  moved image, moved fused labeled image, transformed field grid, and Jacobi deterministic mapping optical flow image.}\label{networkcompair}
			\vspace{-3mm}
		\end{figure*}
		Fig. \ref{networkcompair} displays the registration results of all networks. It is observed that TransFlow is the only network that does not distort or eliminate non-existent labeled images when the labeled images are partially registered. In contrast, all other networks undergo deformation of rigid objects. The optical flow penalty for moving labeled images has minimal effect on the non-existent labeled regions in the fixed images, except for TransFlow. Labeled regions do not play a significant role, while TransFlow can register labeled regions that do not exist in the fixed image but exist in the moving image. Only one-sided labeling is needed to register the CT and MR images. In terms of label fitting, the conventional methods ElasticSync, VoxMorph1, VoxelMorph2, ACRegNet, Vit-V-Net, TransMorph, LKUNet, and PIViT all showed incomplete fitting, which is similar to the evaluation of DSC scores in the Table. ~\ref{compair-net}. It is worth noting that the deformation fields of all the networks do not result in significant folding. We mapped the Jacobian determinant of the optical flow field to a range between 0 and 1 and displayed it, using red and blue to indicate areas of high variability and white to indicate areas of low variability. TransFlow focuses more on the edges and labeled portion of the image, paying less attention to the rest of the image. In contrast, ElasticSyN, VoxelMorph1, VoxelMorph2, PIViT, ACRegNet, Vit-V-Net, TransMorph, and LKUNet exhibit a lack of smoothness in the optical flow field and pay less attention to the regions we want to register. 
		
		Notably, only TransFlow focused on the pubic region. However, the pubic bone was not labeled in all moving images, indicating that the TranFlow network relied solely on weakly supervised learning of other labels to identify the pubic region in moving images. It made subtle adjustments based on the pubic bone's location in fixed images, suggesting that TransFlow prioritizes edge information in images over other networks. It has a certain level of generalization ability.
		\begin{table}[h]
			\caption{Comparison of different registration networks according to DSC,HD,Std(JD),\%|JD|<0}\label{compair-net2}
			\setlength{\tabcolsep}{1.1mm}{
				\begin{tabular}{@{}p{3.25cm}llll}	
					\toprule
					Network& DSC~$\uparrow$ &HD~$\downarrow$&std(JD)~$\downarrow$&\%|JD|<0~$\downarrow$ \\ \midrule
					TransFLow-NoDenosing&0.807&9.254&0.304&0.008\\
					TransFlow-NoOffset&0.771&10.164&0.342&0.008\\
					TransFlow&\textbf{0.825}&\textbf{8.763}&\textbf{0.298}&\textbf{0.006} \\
					\bottomrule
				\end{tabular}
			}
			\vspace{-3mm}
		\end{table}
		
		\begin{figure*}[htbp]
			\centering
			\includegraphics[width=1\linewidth]{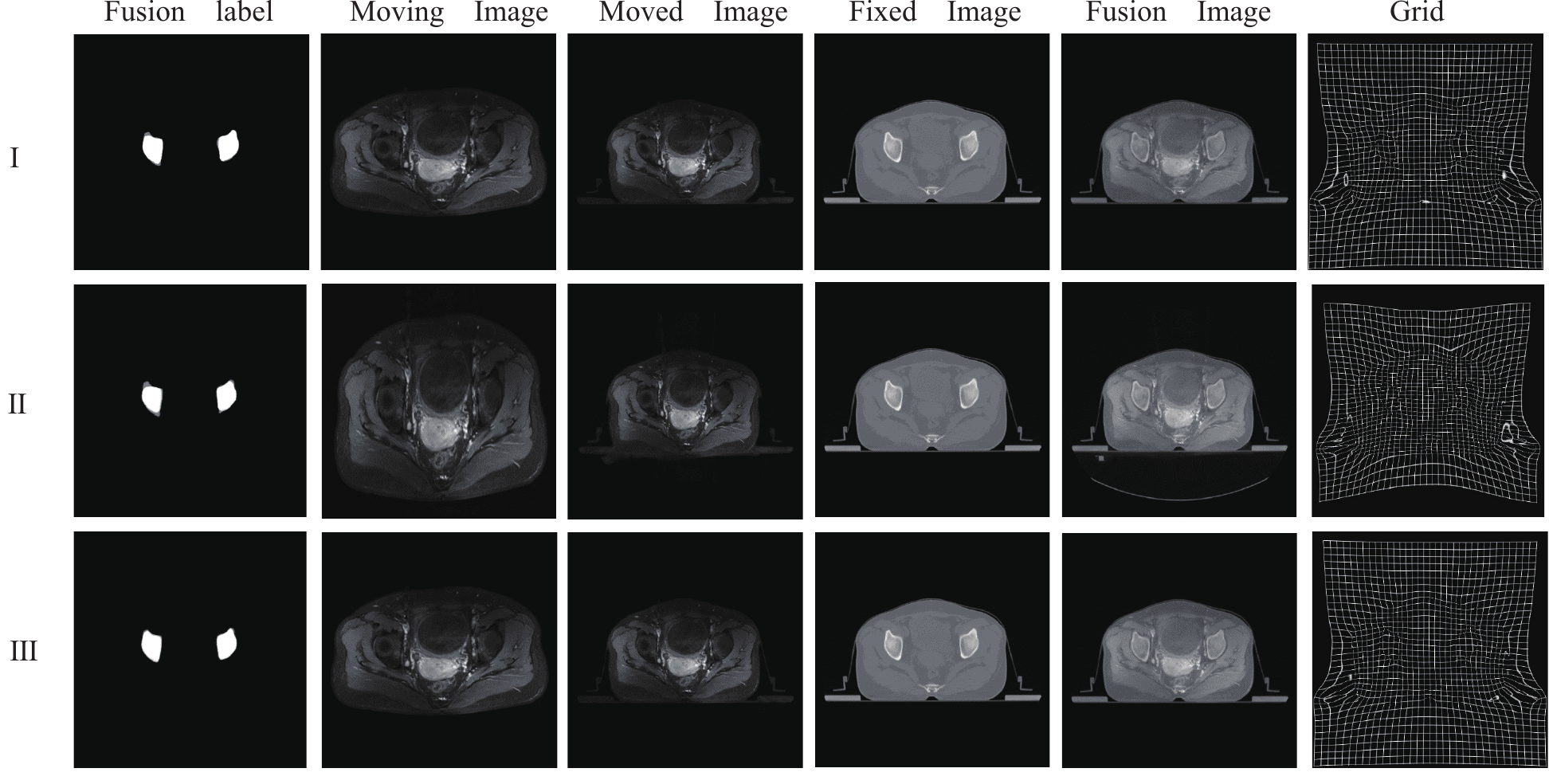}
			\caption{\uppercase\expandafter{\romannumeral1}:TransFlow-NoDenosing \uppercase\expandafter{\romannumeral2}:TransFlow-NoOffset \uppercase\expandafter{\romannumeral3}:TransFlow, from the left to the right columns are Post-registration Fusion Label, Moving Image, Post-registration Image, Fixed Image, Post-registration Fusion Image, Transformation Field Grid.}\label{compair3}
			\vspace{-3mm}
		\end{figure*}
		\begin{figure}[!h]
		\centering
		\includegraphics[width=0.7\linewidth]{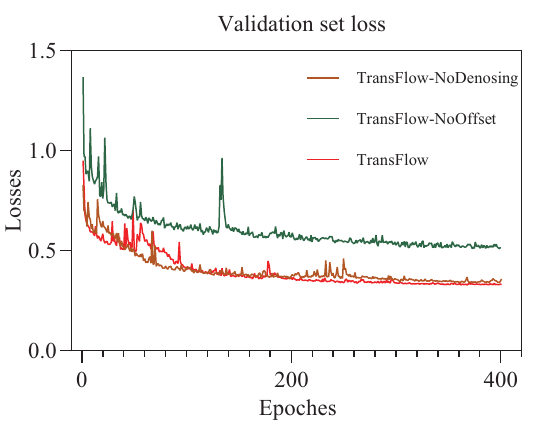}
		\caption{Comparison of Verification Losses for Various Registration Networks}\label{loss}
		\vspace{-3mm}
	\end{figure}
		\subsection{Ablation studies}
		We  compared the effects of using offset correction and denoising operations  on the registration results. Without offset correction and denoising operations, Dice scores decreased by 6.5\%, HD scores increased by 15.9\%, Jacobian determinant variance scores increased by 14.7\%, and non-positive Jacobian determinant percentile scores increased by 33.3\%. And according to the Fig. \ref{compair3}, employing offset correction can significantly minimize optical flow distortion and improve the smoothness of the optical flow field. Additionally, utilizing denoising operations can enhance the registration of rigid labels. We have examined the effects of these three strategies on  network training, 
		According to the Fig. \ref{loss}, we can observe that implementing offset correction significantly enhances the reduction of loss in the validation set. After training 400 times, the loss of the validation set without offset correction is approximately 0.2 higher than the loss with offset correction. We speculate that this difference may be attributed to the absence of offset correction, leaving the moving image in the state depicted in Fig. \ref{compair3} - \uppercase\expandafter{\romannumeral2}, which has implications for the registration network. Whether denoising is performed or not has no significant effect on the training of the network.
		
		\section{Conclusion}
		In this paper, we propose a method for spatial positional alignment of multimodal images. This method can  align multimodal images with existing fixed labels and correct the offset of ROI in the multimodal images. Our alignment method is more efficient and accurate compared to  volume rendering. Based on the alignment, we  propose a weakly supervised learning TransFlow registration network. This network considers pyramidal features with cost volume to estimate the optical flow field of the registration. In comparison with other networks, our method achieves very good performance in several evaluation metrics, proving the effectiveness of our model in multimodal cervical image registration. Our future work involves training a multimodal segmentation network for fully automated alignment and registration. We also aim to optimize the registration network by considering the registration of tissues and organs, and to generalize it to other modalities.
		\bibliographystyle{cas-model2-names}

	\end{sloppypar}
\end{document}